\pdfoutput=1
\documentclass[11pt]{article}
\usepackage[final]{acl}
\usepackage{times}
\usepackage{latexsym}
\usepackage[T1]{fontenc}
\usepackage[utf8]{inputenc}
\usepackage{microtype}
\usepackage{inconsolata}
\usepackage{graphicx}
\usepackage{multirow}

\title{Yes-MT's Submission to the Low-Resource Indic Language Translation Shared Task in WMT 2024}

\author{
    \fontsize{12}{14}\bfseries Yash Bhaskar\textsuperscript{1}, \bfseries Parameswari Krishnamurthy\textsuperscript{2}  \\
    \fontsize{12}{14} IIIT Hyderabad \\
    \texttt{yash.bhaskar@research.iiit.ac.in, param.krishna@iiit.ac.in}
}

\begin{document}
\maketitle
\begin{abstract}
This paper presents the systems submitted by the Yes-MT team for the Low-Resource Indic Language Translation Shared Task at WMT 2024 \cite{pakray-etal-2024-findings}, focusing on translating between English and the Assamese, Mizo, Khasi, and Manipuri languages. The experiments explored various approaches, including fine-tuning pre-trained models like mT5 \cite{xue2021mt5massivelymultilingualpretrained} and IndicBart \cite{Dabre_2022} in both Multilingual and Monolingual settings, LoRA \cite{hu2021lora} finetune IndicTrans2 \cite{gala2023indictrans2}, zero-shot and few-shot prompting \cite{brown2020language} with large language models (LLMs) like Llama 3 \cite{dubey2024llama} and Mixtral 8x7b \cite{jiang2024mixtral}, LoRA Supervised Fine Tuning \cite{mecklenburg2024injecting} Llama 3, and training Transformers \cite{vaswani2017attention} from scratch. The results were evaluated on the WMT23 Low-Resource Indic Language Translation Shared Task's test data using SacreBLEU \cite{post2018call} and CHRF \cite{popovic2015chrf} highlighting the challenges of low-resource translation and show the potential of LLMs for these tasks, particularly with fine-tuning.
\end{abstract}

\section{Introduction}

Developing robust machine translation systems for India's diverse languages is crucial given the country's growing economic importance and the increasing availability of digital content.
However, a significant challenge in developing effective translation tools arises from the limited availability of data for many Indian languages, particularly those spoken in the northeastern regions. This paper describes the Yes-MT team's efforts to address this challenge by participating in the WMT 2024 Low-Resource Indic Language Translation Shared Task, focusing on English to Assamese, Mizo, Khasi, and Manipuri translation. We explored techniques like fine-tuning pre-trained models (mT5, IndicBart) and utilizing large language models (LLMs) like Llama 3 and Mixtral for zero-shot and few-shot learning. Furthermore, we explored using the LoRA technique to fine-tune the IndicTrans2 model, and we also trained Transformer models from scratch. Our findings provide valuable insights into the strengths and weaknesses of different approaches, highlighting the potential of LLMs and fine-tuning techniques in overcoming the limitations of data scarcity.

\section{Dataset}

The dataset used in this study consists of parallel bilingual data provided by the WMT 2024 Low-Resource Indic Language Translation Shared Task organizers \cite{pal2023findings} \& \cite{pakray-etal-2024-findings}. The training, validation, and test splits for each language pair are detailed in Table 1.

\begin{table}[htbp]
\centering
\begin{tabular}{|c|c|c|c|}
\hline
\textbf{Language Pair} & \textbf{Train} & \textbf{Val} & \textbf{Test} \\ \hline
Assamese (en-as) & 50,000 & 2,000 & 2,000 \\ \hline
Mizo (en-lus) & 50,000 & 1,500 & 2,000 \\ \hline
Khasi (en-kha) & 24,000 & 1,000 & 1,000 \\ \hline
Manipuri (en-mni) & 21,000 & 1,000 & 1,000 \\ \hline
\end{tabular}
\caption{Number of Sentences in Train, Validation, and Test Sets}
\label{tab}
\end{table}

In addition to the bilingual data, we also had access to a significant amount of Monolingual data for each of the target languages, which included 2.60 million sentences in Assamese, 1.90 million sentences in Mizo, 0.18 million sentences in Khasi, and 2.10 million sentences in Manipuri. However, for the scope of this work, we focused exclusively on utilizing the provided bilingual data for training and evaluation, aiming to explore the capabilities of the models under truly low-resource conditions.

Limiting our study to the provided bilingual data allowed us to maintain a consistent and controlled experimental environment, ensuring the results reflected the performance of our approaches under the typical constraints of low-resource language translation tasks. In the future, we may explore incorporating the available monolingual data, such as through back-translation, to further improve translation quality.

\section{Experiments}

This section details the experimental setup used for the various models and training strategies employed in our submission.

\subsection{Primary Submission}

Our primary submission involved training a Transformer model from scratch using the Fairseq framework \cite{ott-etal-2019-fairseq}. This model was trained for Multilingual translation, handling all four language directions (English to Assamese, Manipuri, Mizo, and Khasi) simultaneously. We utilized BPE tokenizer \cite{araabi2022effective} and Transformer architecture. The architectural details are shown in Table 2.

\begin{table}[h]
\centering
\begin{tabular}{|c|c|}
\hline
\textbf{Parameter} & \textbf{Value} \\ \hline
Embedding Dimension & 512 \\ \hline
FFN Dimension & 1024 \\ \hline
Attention Heads & 4 \\ \hline
Encoder Layers & 6 \\ \hline
Decoder Layers & 6 \\ \hline
\end{tabular}
\caption{Transformer Architecture Details}
\end{table}

\subsection{Contrastive Submission}

The contrastive submission explored fine-tuning pre-trained models in two settings: language-specific and Multilingual.

\subsubsection{Multilingual Fine-tuning:} Both mT5 and IndicBart were fine-tuned in a Multilingual setting, where a single model was trained to handle all four language directions. To enable the models to distinguish between the target languages, we added language-specific tokens to their existing vocabularies, as suggested by previous work \cite{johnson2017google}. The language-specific tokens used are shown in Table 3. A single model was trained for one-to-many translation across all four language directions for each of the indicBart, mT5-small, and IndicTrans2 systems. The results are in Table 4. IndicBart and mT5-small were fine-tuned using Full Fine-Tuning (FFT), while IndicTrans2 was fine-tuned employing the LoRA (Low-Rank Adaptation) technique \cite{hu2021lora}.

\begin{table}[h]
\centering
\begin{tabular}{|c|c|}
\hline
\textbf{Language} & \textbf{Token} \\ \hline
Assamese (asm) & `<asm\_Beng>` \\ \hline
Manipuri (mni) & `<mni\_Beng>` \\ \hline
Khasi (kha) & `<kha\_Latn>` \\ \hline
Mizo (lus) & `<lus\_Latn>` \\ \hline
\end{tabular}
\caption{Language-Specific Tokens}
\end{table}

\subsubsection{Monolingual Fine-tuning:} We also trained separate models for each language pair, as these focused on a single translation direction and did not require language-specific tokens.

For each language direction, we trained four distinct models using mT5-Small and IndicBart with Full Fine-Tuning (FFT). The results are in Table 4.

\begin{table*}[htbp]
\centering
\begin{tabular}{|c|c|c|c|c|c|}
\hline
\textbf{Model} & \textbf{Training Type} & \textbf{en-as} & \textbf{en-kha} & \textbf{en-mz} & \textbf{en-mni} \\ \hline
Transformers & Multilingual & 16.06 & 19.67 & 5.49 & 20.60 \\ \hline
\multirow{2}{*}{IndicBart} 
                     & Monolingual & 6.4 & 11.2 & 25.1 & 8.8 \\ \cline{2-6}
                     & Multilingual & 6.5 & 11.4 & 25.3 & 9.1 \\ \hline
\multirow{2}{*}{mT5-small} 
                     & Monolingual & 14.3 & 12.9 & 31.4 & 19.2 \\ \cline{2-6}
                     & Multilingual & 15.6 & 13.6 & 32.3 & 23.9 \\ \hline
IndicTrans2-2B
                     & ZeroShot & 49.2 & - & - & 44.9 \\ \hline
\multirow{2}{*}{IndicTrans2-200M} 
                     & ZeroShot & 49.5 & - & - & 45.3 \\ \cline{2-6}
                     & Multilingual & 47.27 & - & - & 49.12 \\ \hline
\end{tabular}
\caption{ChrF Scores for 
Monolingual : Models fine-tuned for one-to-one language translation \\
Multilingual : Models fine-tuned for one-to-many language translation}
\label{tab:chrf_scores}

\end{table*}

\subsection{Experiments with LLMs}

Additionally, we explored the use of the Llama3 model in conjunction with the LoRA (Low-Rank Adaptation) technique.

\paragraph{Zero-Shot and Few-Shot Translation Evaluation} We tested Zero Shot Translation capabilities of Llama 3-8B-8192, Llama 3-70B-8192, mixtral-8x7B-32768, Llama3-8B-instruct and Llama3.1-8B-instruct. We also tested the few-shot translation capabilities of Llama3.1-8B-instruct with 3-shot, 5-shot, and 10-shot prompting.

\paragraph{Supervised Fine-Tuning with LoRA} We fine-tuned a 4-bit quantized \cite{liu2023llm} Llama3 model using the LoRA technique with Supervised Fine-Tuning (SFT), employing the LlamaFactory framework \cite{zheng2024llamafactory}. We used a prompt-based approach for translation, providing the model with a system prompt and a prompt template specifying the source and target languages.

The following template was used for fine-tuning the Large Language Models (LLMs):
\begin{verbatim}
System Prompt : You are a helpful assistant.
Prompt Template : Translate the following
English sentence to {target_language} in
{target_script} Script:\n{input_sent}
\end{verbatim}

\section{Results}

\subsection{Multilingual vs. Monolingual Performance}

One key finding from our experiments was the performance comparison between the Multilingual and Monolingual training approaches for the mT5 and IndicBart models. As shown in Table 4, the Multilingual versions of both mT5 and IndicBart consistently outperformed their Monolingual counterparts across the translation tasks.

\begin{itemize}
\item For mT5, the Multilingual model outperformed the Monolingual model across all language pairs, with ChrF score improvements ranging from 1.3 to 4.7 points. This suggests that mT5 benefits from the shared linguistic knowledge across different languages in a Multilingual setting, which enhances its ability to generalize to low-resource languages.
\item Likewise, IndicBart demonstrated a slight performance boost in the Multilingual setting compared to the Monolingual models, suggesting that the Multilingual training approach provided a benefit.
\end{itemize}

The better performance of the Multilingual models is likely due to the shared linguistic knowledge they gained during training, which may have provided a richer context and improved their ability to generalize. This indicates that leveraging Multilingual data, even in limited-resource scenarios, can be a more effective approach than focusing on Monolingual training.

\begin{table*}[htbp]
\centering
\begin{tabular}{|c|c|c|c|c|c|}
\hline
\textbf{Model} & \textbf{Inference} & \textbf{en-as} & \textbf{en-kha} & \textbf{en-mz} & \textbf{en-mni} \\ \hline

Llama3-8B-8192        & Zero Shot & 18.56 & 14.92 & 15.57 & 13.45 \\ \hline
Llama3-70B-8192       & Zero Shot & 27.54 & 18.57 & 20.62 & 15.53 \\ \hline
mixtral-8x7B-32768    & Zero Shot & 6.79  & 15.45 & 16.57 & 2.65  \\ \hline
\multirow{3}{*}{Llama3-8B-instruct} 
                     & Zero Shot  & 26.13 & 8.38  & 18.06 & 15.29 \\ \cline{2-6}
                     & 1 Epoch    & 29.82 & 33.19  & 32.72 & 37.85 \\ \cline{2-6}
                     & 2 Epoch    & 31.68 & 35.26  & 37.73 & 44.51 \\ \hline
\multirow{4}{*}{Llama3.1-8B-instruct} 
                     & Zero Shot & 22.93 & 12.03 & 15.23 & 14.47 \\ \cline{2-6}
                     & 3 Shot    & 23.26 & 13.66 & 18.89 & 15.30 \\ \cline{2-6}
                     & 5 Shot    & 23.48 & 15.11 & 18.77 & 15.29 \\ \cline{2-6}
                     & 10 Shot   & 23.89 & 16.03 & 19.39 & 15.43 \\ \hline

\end{tabular}
\caption{ChrF Scores for Various Models, Shot Types, and Language Pairs}
\label{tab:chrf_scores_llm}
\end{table*}

\subsection{Expected Structured Output}

A challenge observed during the experiments was the generation of structured output. Ideally, the output should directly provide the translated sentence without additional, unnecessary text. However, we noticed that the LLM models sometimes wrapped the translation in extraneous text, such as “The translation of the given sentence is: {Translation}”, followed by further analysis and explaination making it difficult to extract the translation. This adds noise to the output and complicates the process of extracting the actual translation.

We analyzed the percentage of outputs that were wrapped with unnecessary text across different settings:

This issue of unnecessary text in the output was more common in the zero-shot setting, where 66.\% of the outputs included additional text. As the number of shots increased, the percentage of such outputs decreased significantly to 0.18\% in 10 Shot Prompting, indicating that few-shot prompting can help guide the LLM to produce more structured and concise translations.

To improve the usability of LLM-based machine translation systems, it's crucial to fine-tune the models or design prompts that consistently yield clean and structured outputs, particularly in low-resource settings where post-processing resources might be limited.

\begin{figure}
    \centering
    \includegraphics[width=1\linewidth]{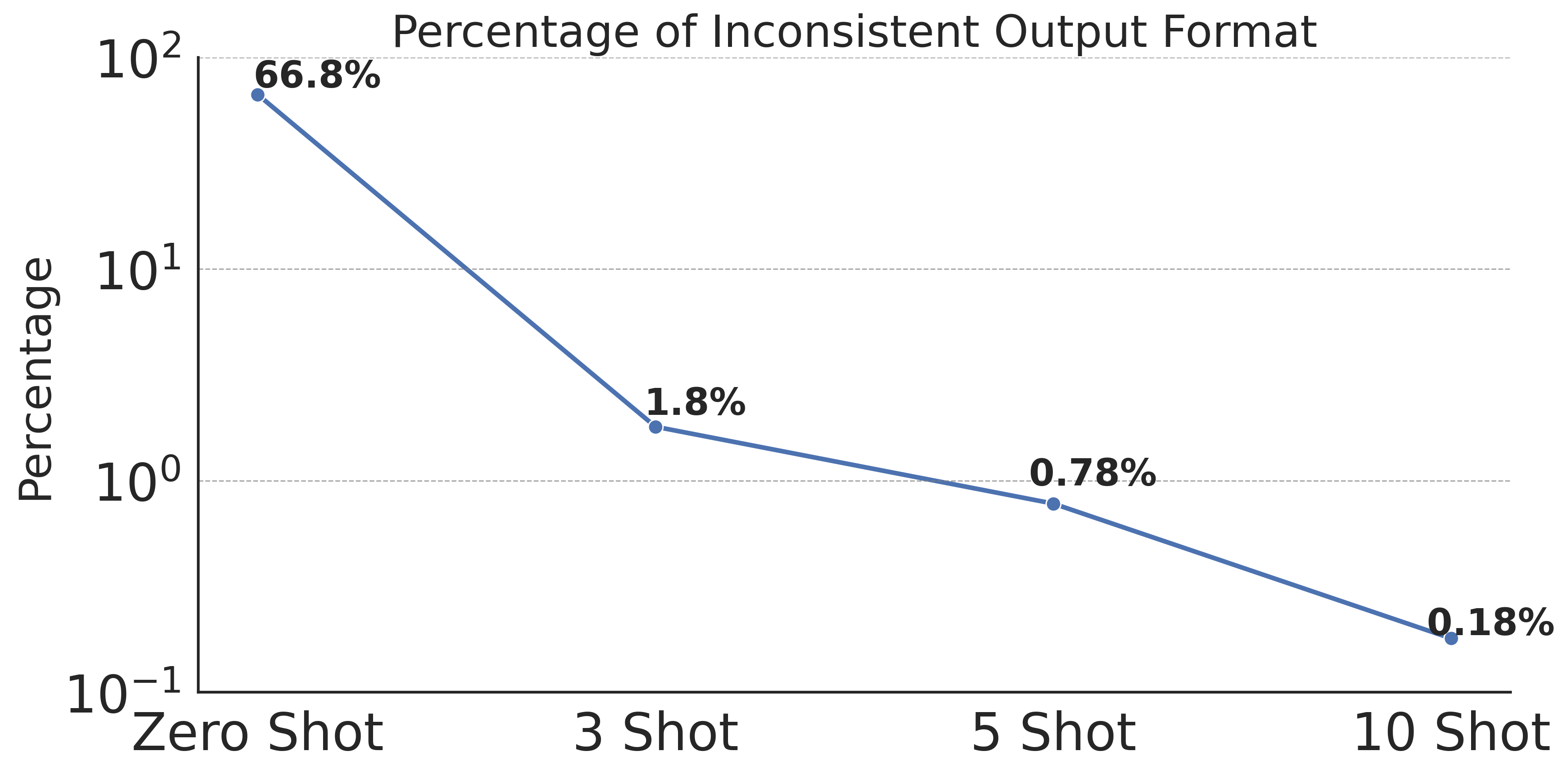}
    \caption{Inconsistent Output Format with Few Shot Prompting}
    \label{fig:structured-output}
\end{figure}

\section{WMT 2024 Results}

The performance of our models on the WMT 2024 Low-Resource Indic Language Translation Shared Task dataset is summarized in the following table, focusing on the ChrF \cite{popovic2015chrf} metrics:
\begin{table}[htbp]
\centering
\begin{tabular}{|c|c|c|}
\hline
\textbf{Language Pair} & \textbf{Submission Type} & \textbf{ChrF} \\ \hline
Eng-Asm & primary & 0.1123 \\ 
         & contrastive & 0.6518 \\ \hline
Eng-Mni & primary & 0.1102 \\ 
         & contrastive & 0.4438 \\ \hline
Eng-Lus & primary & 0.1282 \\ 
        & contrastive & 0.4151 \\ \hline
Eng-Kha & primary & 0.1139 \\ 
        & contrastive & 0.3541 \\ \hline
\end{tabular}
\caption{ChrF Scores for WMT 2024 Shared Task}
\label{tab:wmt2024_chrf_scores}
\end{table}

For the primary submissions, we utilized Transformers trained from scratch without additional data. As indicated by the scores, the primary systems struggled significantly, yielding very low ChrF values across all language pairs.

In contrast, the models fine-tuned for the contrastive submissions demonstrated noticeable improvements. For Assamese and Manipuri, we fine-tuned IndicTrans2, achieving the highest ChrF scores in these language pairs. For Mizo and Khasi, we fine-tuned Llama3, which also resulted in enhanced performance compared to the primary systems. These findings highlight the effectiveness of fine-tuning pre-trained models, even in low-resource settings.

\section{Potential Test Set Bias}

One of the noteworthy observations in this year (2024) WMT 2024 results is the significant difference in the performance of the primary Transformers trained from scratch when evaluated on this year's (2024) test set compared to last year's (2023) test set. Specifically, we observed that the models performed better on last year's test set despite using the same training data.

This discrepancy could be indicative of a translation bias present in last year's dataset, which might have inadvertently favored the models trained on that data. The primary systems, having been trained exclusively on the previous year's data, may have overfitted to patterns specific to that dataset, leading to better performance on the older test set but struggling on the newer one.

This implies that the primary models may have difficulty generalizing to entirely new data distributions, an important factor to consider in low-resource settings where the training data is limited and may not be representative of future data. It also underscores the importance of using diverse and varied datasets during training to help mitigate such biases and improve the overall robustness of the models.

\section{Conclusion}

This paper presented the systems and results of the Yes-MT team's participation in the WMT 2024 Low-Resource Indic Language Translation Shared Task. The experiments highlighted the potential of LLMs, especially when fine-tuned with techniques such as LoRA, in enhancing translation quality even under low-resource conditions. The contrastive submissions, which utilized fine-tuned LLMs, demonstrated significant improvements over the primary submissions that relied on training Transformers from scratch.

Our findings suggest that while training models from scratch can be challenging in low-resource settings due to data scarcity and generalization issues, fine-tuning pre-trained models can effectively bridge the gap, leveraging shared knowledge across languages to achieve better translation performance.

Future work could explore integrating monolingual data through back-translation or other data augmentation techniques, as well as further refining prompt engineering strategies to improve the structure and clarity of LLM outputs. Additionally, focusing on addressing potential biases in test data to help create more reliable translation systems.

\section*{Acknowledgments}

We acknowledge the organizers of the WMT 2024 Low-Resource Indic Language Translation Shared Task for providing the valuable dataset and facilitating this research. We also thank the developers of the mT5, IndicBart, IndicTrans2, Llama 3, and Mixtral models for making their work publicly available.

\bibliography{custom}

\end{document}